\documentclass[final,5p,times,twocolumn]{elsarticle}

\usepackage{amsmath,amssymb,amsfonts}
\usepackage{xcolor}
\usepackage{enumitem}
\usepackage{graphicx}
\usepackage{booktabs}
\usepackage{algorithm}
\usepackage{algpseudocode}
\usepackage{orcidlink}

\newcommand{\gup}[1]{\textcolor{blue}{$\uparrow$\,#1}}
\newcommand{\gdn}[1]{\textcolor{red}{$\downarrow$\,#1}}

\journal{Journal of Manufacturing Systems}

\begin{document}

\begin{frontmatter}

\title{From Comparison to Selection: Enhancing Cross-Scale Generalization of Pointer Networks Trained with Reinforcement Learning for Capacitated Vehicle Routing Problems}

\author[inst1]{Sofiene Lassoued \orcidlink{0000-0001-7919-6939}*}
\author[inst1]{Devara Tertia}
\author[inst2]{Stefan Lier      \orcidlink{0000-0002-3314-7610}}
\author[inst1]{Andreas Schwung  \orcidlink{0000-0001-8405-0977}}

\cortext[cor1]{Corresponding author: Sofiene Lassoued, email: lassoued.sofiene@fh-swf.de}

\affiliation[inst1]{organization={South Westphalia University of Applied Sciences},
                    department={Automation Technology and Learning Systems},
                    addressline={Luebecker Ring 2},
                    city={Soest},
                    postcode={59494},
                    state={North Rhine-Westphalia},
                    country={Germany}}

\affiliation[inst2]{organization={South Westphalia University of Applied Sciences},
                    department={Logistik und Supply Chain Management},
                    addressline={Lindenstr.53},
                    city={Meschede},
                    postcode={59872},
                    state={North Rhine-Westphalia},
                    country={Germany}}

\begin{abstract}
Vehicle routing in delivery and logistics remains challenging as the number of clients varies. Classical approaches based on mixed-integer programming and heuristics can scale poorly or require problem-specific design, motivating learning-based alternatives. Reinforcement learning is particularly promising, provided a single policy can generalise across problem sizes without retraining. Although attention-based Pointer Networks are inherently size-agnostic, their solution quality degrades under both upscaling and downscaling. Using the Capacitated Vehicle Routing Problem (CVRP) we investigate this degradation and identify three contributing factors: shifts in node density, attention dispersion, and dilution of the mean-pooled graph context. We systematically compare architecture-wide, encoder-side, and decoder-side modifications to address these limitations and propose a learned, permutation-invariant graph-context aggregation mechanism, implemented as a weighted average and as Set2Set. Results reveal distinct requirements for transfer in each direction: entropy-based attention scaling reduces the CVRP20 to CVRP100 performance gap by 1.41 percentage points, while the proposed aggregation methods achieve the largest improvement on the more challenging CVRP100 to CVRP20 transfer, approximately 0.9 percentage points. Exploiting this complementarity, we introduce an approach selector that chooses the best model for each training–target size pair. The selector reduces cost relative to the reference model across all nine seen train–test pairs and eight of nine pairs involving unseen sizes (CVRP35, CVRP75, and CVRP200), achieving improvements of up to 3.2\%.

\end{abstract}


\end{frontmatter}

\section{Introduction}
\label{sec:intro}

The growth of e-commerce, on-demand services, and urban logistics pushes delivery operators to plan routes under tight constraints on capacity, distance, energy, and cost. Delivery fleets can consist of ground vehicles as well as drones, among others~\cite{ostermann.2025a}. This work investigates learning-based route planning under a vehicle capacity constraint, independent of the vehicle type. From an optimization standpoint, the task is the Capacitated Vehicle Routing Problem (CVRP): a fleet must serve customers with known demands within a capacity limit at minimal total route length~\cite{dantzig1959,toth2014}. CVRP isolates the capacity-constrained routing core and is a standard controlled benchmark for routing algorithms, including learning-based ones~\cite{nazari2018,kool2019}.

Exact methods scale poorly on this NP-hard problem~\cite{fukasawa2006,pessoa2020}, and heuristics need problem-specific, hand-tuned design~\cite{laporte2002,vidal2013}. Neural Combinatorial Optimization (NCO) instead trains a model to construct solutions directly, using reinforcement learning (RL) so that no optimal-solution labels, which are expensive for large CVRP instances, are needed~\cite{bello2016,kool2019,bengio2021}.

The central difficulty for route-planning policy is that the number of clients is not fixed: it varies from day to day and region to region. Training one policy per client count is not viable, so the policy must construct solutions for a variable, a priori unknown number of nodes, which is a known weakness of neural routing solvers~\cite{manchanda2022}. A prominent size-agnostic approach is the Pointer Network with attention, which points to one of the input nodes at each decoding step instead of choosing from a fixed vocabulary, so the same trained policy can be applied to graphs of any size~\cite{vinyals2015,kool2019}. Trained with RL, it has been benchmarked at scale in recent studies of neural routing solvers~\cite{liu2023a,berto2023}, which show that it solves instances of the training size well but that generalization across problem sizes remains a clear bottleneck in both directions: performance degrades when the model is \emph{upscaled} to larger graphs and when it is \emph{downscaled} to smaller ones.

This paper therefore investigates which modifications of the attention-based Pointer Network, applied architecture-wide, in the encoder or in the decoder, reduce the upscaling and the downscaling generalization gap for CVRP, and how their complementary strengths can be combined in one size-agnostic routing solver.

The study focuses on the encoder--decoder architecture of the attention-based Pointer Network: the encoder embeds the nodes of an instance, and the decoder uses these embeddings to point to the next node at each step. This architecture is kept as the common frame, and every approach modifies it at one of three levels: architecture-wide, acting on the attention of both encoder and decoder; encoder-side, replacing how node embeddings are computed; or decoder-side, changing how the decoder summarises the instance and builds its query. To this end, we first identify the sources of the cross-size degradation of the reference policy, then assign to each source the levers that act on it, and finally compare all levers head-to-head under an identical training and evaluation protocol, scored by how much they reduce the cross-size optimality gap of the reference in each direction.

\paragraph{Contributions} 

The contributions of this paper are:
\begin{enumerate}[label=(\roman*),leftmargin=*,itemsep=1pt,topsep=2pt]
\item We conduct a diagnosis-driven, head-to-head comparison of seven levers at the architecture-wide, encoder and decoder levels of the Pointer Network, with fixed vehicle capacity so that only the number of clients changes. It shows that upscaling and downscaling respond to different levers.\item Based on the comparison, we propose a learned, permutation-invariant graph-context aggregation for the attention-model decoder, realised as a weighted average with a scoring network and as a Set2Set summary. It replaces the mean-pooled context and gives the largest gain of all evaluated approaches on the hardest downscaling transfer while keeping same-size quality.
\item We further propose an approach selector that exploits this complementarity by choosing, for a given training size and target size, the approach with the lowest estimated cost. It lowers the cost of the reference on seen and unseen problem sizes.
\end{enumerate}

This paper is structured as follows: Section~\ref{sec:related} reviews related work, and Section~\ref{sec:background} introduces the necessary background on graph neural networks, attention and pointer decoding. Section~\ref{sec:formulation} formulates the CVRP as a sequential decision problem, gives the training objective and describes the reference policy. Section~\ref{sec:method} derives the sources of cross-size degradation and presents the evaluated approaches together with the proposed graph-context aggregation and approach selector. Section~\ref{sec:setup} details the experimental setup, Section~\ref{sec:results} reports and discusses the results, and Section~\ref{sec:conclusion} concludes the paper and outlines future work.

\section{Related Work}
\label{sec:related}

\subsection{Combinatorial optimization and routing}
Combinatorial optimization (CO) covers problems with a combinatorial explosion of feasible solutions~\cite{korte2018}. Routing problems, notably the Traveling Salesman Problem and the Vehicle Routing Problem, are among the most studied because of their relevance to logistics~\cite{toth2014,laporte2009}. The Capacitated VRP (CVRP) adds a capacity constraint that partitions customers across routes~\cite{dantzig1959,clarke1964}. CVRP has been addressed with exact methods~\cite{augerat1995,fukasawa2006}, heuristics such as the savings algorithm and 2-opt/3-opt~\cite{clarke1964,lin1965}, and metaheuristics such as tabu search and hybrid genetic search~\cite{gendreau1994,vidal2013}. Nevertheless, exact methods scale poorly and heuristics need problem-specific tuning~\cite{toth2014,laporte2002}.

\subsection{Machine learning for combinatorial optimization}
Machine learning has assisted classical solvers by learning branching decisions~\cite{khalil2016,gasse2019}, cutting-plane selection~\cite{tang2020,huang2022}, solver configuration and algorithm selection~\cite{hutter2011,kerschke2019}, and warm starts~\cite{nair2020}, but the solver itself remains classical. NCO instead uses a neural network as the solver, learning to construct a solution end to end~\cite{bengio2021,mazyavkina2021}.

Early NCO approaches for routing used size-agnostic Pointer Networks~\cite{vinyals2015} (Section~\ref{sec:background}). Bello et al.\ trained a Pointer-Network policy with actor-critic RL for TSP~\cite{bello2016}, and Nazari et al.\ extended it to CVRP with capacity-aware masking~\cite{nazari2018}. Kool et al.\ replaced the recurrent encoder with a Transformer-style multi-head-attention (MHA) encoder and trained the resulting attention-model policy with REINFORCE and a rollout baseline, with strong results on TSP, CVRP, the Orienteering Problem, and the Prize-Collecting TSP~\cite{kool2019,vaswani2017}. This architecture underlies RL4CO~\cite{berto2023} which is the reference architecture used throughout this paper.

\subsection{The size-generalization bottleneck in attention-based NCO}
A systematic evaluation of neural routing solvers shows that the attention-based Pointer-Network policy performs well on instances matching its training size, but not when the graph size changes at test time~\cite{liu2023a}, as also reported by RL4CO~\cite{berto2023}. The degradation is bidirectional: models trained on small graphs lose quality when \emph{upscaled}, and models trained on large graphs do not transfer cleanly when \emph{downscaled}. NCO models more broadly fail on unseen sizes~\cite{manchanda2022}. Meta-learning~\cite{manchanda2022} and test-time adaptation~\cite{chen2025} address varying task distributions, including size variation. At the level of the attention computation, Wang et al.\ attribute part of the degradation to attention dispersion as the number of nodes grows~\cite{wang2025}, and Xiao et al.\ counteract it with an entropy-based scaling factor applied to the attention scores at inference time~\cite{xiao2025}. Architectural work asks whether the encoder or decoder composition can be changed: Berto et al.\ show that the encoder choice affects size generalization~\cite{berto2023}, and Luo et al.\ propose a Light-Encoder-Heavy-Decoder (LEHD) composition, in contrast to the Heavy-Encoder-Light-Decoder design of Kool et al., with substantially improved generalization under supervised learning~\cite{luo2023}. The Set Transformer~\cite{lee2019} offers a further route to size-independent encoding through a fixed number of inducing points.

\subsection{Research gap}
Most studies target a single component in isolation and evaluate same-size performance or one transfer direction. In addition, attention-scaling methods have been evaluated in settings where vehicle capacity grows with the problem size, which confounds the effect of the number of nodes with that of the demand regime. The graph-level context that the decoder reuses at every step, a mean over node embeddings in the attention model, has received less attention as a lever for size generalization. Finally, algorithm selection is established for classical solvers~\cite{kerschke2019}, but, to our knowledge, it has not been used to combine NCO variants that generalize in different directions. This paper closes these gaps with a learned graph-context aggregation, a controlled comparison of levers in both directions, and an approach selector built on that comparison.

\section{Theoretical Background}
\label{sec:background}

We start by presenting the theoretical background focusing on the neural network architectures employed for RL-driven CVRP solutions in this paper.

\subsection{Node embeddings and graph neural networks}
Neural routing policies do not operate on raw node features directly. An encoder first maps the features $x_v$ of every node $v$ of a graph $G=(V,E)$ to an embedding $h_v\in\mathbb{R}^{D}$ that also reflects the node's relation to the rest of the graph. Because the embedding dimension $D$ is fixed, the same downstream layers can process graphs with any number of nodes, which is the basis of size-agnostic policies. A graph-level representation is then obtained with a permutation-invariant readout, most simply the mean of all node embeddings.

Graph neural networks (GNNs) compute node embeddings by message passing~\cite{scarselli2009,gilmer2017}. In each layer $k$, every node aggregates the embeddings of its neighbours $N(v)$ and combines them with its own,
\begin{equation}
h_v^{(k)}=\phi^{(k)}\Big(h_v^{(k-1)},\;\mathrm{AGG}^{(k)}\big(\{h_u^{(k-1)}:u\in N(v)\}\big)\Big),
\label{eq:mp}
\end{equation}
so that after $K$ layers the embedding summarises the $K$-hop neighbourhood of the node. Graph convolutional networks (GCNs)~\cite{kipf2017} realise Eq.~\eqref{eq:mp} with a fixed aggregation weighted by the normalised adjacency matrix,
\begin{equation}
H^{(k)}=\sigma\big(\tilde D^{-1/2}\tilde A\,\tilde D^{-1/2}H^{(k-1)}W^{(k)}\big),
\label{eq:gcn}
\end{equation}
where $\tilde A=A+I$ is the adjacency matrix with self-loops, $\tilde D$ its degree matrix and $W^{(k)}$ a learned weight matrix. Since the propagation matrix is built from one specific graph, GCNs were introduced for learning on a fixed graph.

GraphSAGE~\cite{hamilton2017} instead learns how to combine the embeddings of a node's neighbours, so that the learned rule can be applied to nodes and graphs that were not seen during training. With the mean aggregator, one layer reads
\begin{equation}
\begin{aligned}
h_{N(v)}^{(k)}&=\frac{1}{|N(v)|}\sum_{u\in N(v)}h_u^{(k-1)},\\
h_v^{(k)}&=\sigma\Big(W^{(k)}\big[h_v^{(k-1)}\,\|\,h_{N(v)}^{(k)}\big]\Big),
\end{aligned}
\label{eq:sage}
\end{equation}
where $\|$ denotes concatenation. To bound the computation, the original method uses only a random sample of fixed size from each neighbourhood. Each layer mixes a node's embedding with those of its neighbours, so when many layers are stacked, the embeddings of all nodes become nearly identical and the nodes can no longer be distinguished. This effect, known as over-smoothing~\cite{li2018}, is why GNN encoders use only a few layers.

\subsection{Multi-head attention and attention scaling}
Attention was introduced for sequence alignment in neural machine translation~\cite{bahdanau2014} and generalized to Multi-Head Attention (MHA) in the Transformer~\cite{vaswani2017}. From node embeddings $X$, MHA computes queries $Q=XW^Q$, keys $K=XW^K$ and values $V=XW^V$ and applies scaled dot-product attention,
\begin{equation}
\mathrm{Attention}(Q,K,V) = \mathrm{softmax}\!\left(\frac{QK^\top}{\sqrt{d_k}}\right)V,
\label{eq:attention}
\end{equation}
in $h$ parallel heads whose outputs are concatenated,
\begin{equation}
\mathrm{MHA}(X)=\mathrm{Concat}(\mathrm{head}_1,\dots,\mathrm{head}_h)\,W^O.
\label{eq:mha}
\end{equation}
In self-attention, every node is updated with information from every other node at $O(N^2)$ cost. Because the softmax in Eq.~\eqref{eq:attention} normalises over all $N$ keys, the sharpness of the attention distribution depends on $N$. Xiao et al.~\cite{xiao2025} therefore multiply the attention scores by a size-dependent entropy-based scaling factor (ESF),
\begin{equation}
\begin{aligned}
\mathrm{Attention}_{\lambda}(Q,K,V) &= \mathrm{softmax}\!\left(\lambda_n\frac{QK^\top}{\sqrt{d_k}}\right)V,\\
\lambda_n &= \log_{n_{\mathrm{train}}} n = \frac{\log n}{\log n_{\mathrm{train}}},
\end{aligned}
\label{eq:esf}
\end{equation}
where $n$ is the size of the current instance and $n_{\mathrm{train}}$ the training size. Since $\lambda_n=1$ for $n=n_{\mathrm{train}}$, the factor leaves the trained model unchanged on its training size and can be applied at inference time.

\subsection{Attention on sets: Set Transformer and Set2Set}
A set of nodes has no natural order, so models that process it should not depend on the order in which the nodes are presented. The Set Transformer~\cite{lee2019} builds such a model from multi-head attention blocks (MAB). Its Induced Set Attention Block (ISAB) routes all interaction between the $n$ elements through $m$ learnable inducing points $I\in\mathbb{R}^{m\times d}$,
\begin{equation}
\begin{aligned}
\mathrm{MAB}(X,Y) &= \mathrm{LN}\big(H + \mathrm{rFF}(H)\big), \\
H &= \mathrm{LN}\big(X + \mathrm{Multihead}(X,Y,Y)\big), \\
\mathrm{ISAB}_m(X) &= \mathrm{MAB}\big(X,\mathrm{MAB}(I,X)\big),
\end{aligned}
\label{eq:isab}
\end{equation}
where $\mathrm{LN}$ is layer normalization, $\mathrm{rFF}$ a row-wise feed-forward network, and $X$ and $Y$ provide the queries and the keys/values, respectively. The inner block summarises the $n$ elements into $m$ vectors and the outer block lets every element attend to that summary, an $n\!\to\! m\!\to\! n$ bottleneck whose width does not depend on $n$ and whose cost is $O(nm)$ instead of $O(n^2)$.

Set2Set~\cite{vinyals2016} is an order-invariant readout that summarises a set of embeddings $\{h_i\}$ into one vector. Over $T$ processing steps, an LSTM state attends over the set and reads out a weighted sum,
\begin{equation}
\begin{aligned}
q_t&=\mathrm{LSTM}(q^*_{t-1}), & \alpha_{i,t}&=\mathrm{softmax}_i\big(h_i^\top q_t\big),\\
r_t&=\textstyle\sum_i\alpha_{i,t}h_i, & q^*_t&=[q_t\,\|\,r_t],
\end{aligned}
\label{eq:set2set}
\end{equation}
where $q^*_T$ is the summary. Because each step attends again with an updated query, the readout can capture more than a single average.

\subsection{Route decoding: pointer attention}
Routing outputs are permutations of a variable-size input; the Pointer Network~\cite{vinyals2015} therefore uses attention as a pointer over input positions,
\begin{equation}
p(\pi_t = i \mid \pi_{<t}, X) = \frac{\exp(\tilde{u}^t_i)}{\sum_{j \in \mathcal{A}_t}\exp(\tilde{u}^t_j)}, \;
\tilde{u}^t_i = \begin{cases} u^t_i, & i \in \mathcal{A}_t \\ -\infty, & \text{else}, \end{cases}
\label{eq:pointer}
\end{equation}
where $u^t_i$ is the attention score of input $i$ at step $t$ and $\mathcal{A}_t$ the set of inputs that may still be selected; all other inputs are masked. Because the output distribution is defined over whatever inputs are present, the same model applies to inputs of any size. The attention-model policy of Kool et al.~\cite{kool2019} combines an MHA encoder with such a pointer decoder (Section~\ref{sec:reference}).

\section{Problem Formulation and Reference Policy}
\label{sec:formulation}

\subsection{CVRP as a sequential decision problem}
A CVRP instance is a weighted graph $G=(V,E)$ with $V=\{0,1,\dots,n\}$, where node $0$ is the depot and nodes $1,\dots,n$ are customers with demands $d_i$. The cost between two nodes is their Euclidean distance $c_{ij} = \sqrt{(x_i-x_j)^2 + (y_i-y_j)^2}$. Each vehicle has capacity $Q$, and every route $r_k$ must satisfy $\sum_{i \in r_k} d_i \le Q$. A feasible solution $S=\{r_1,\dots,r_m\}$ visits every customer exactly once, with every route starting and ending at the depot,and minimizes the total cost

\begin{equation}
C(S)=\sum_{k=1}^m C(r_k), \qquad C(r_k)=c_{0i_1}+\sum_{t=1}^{L_k-1}c_{i_t i_{t+1}}+c_{i_{L_k}0},
\label{eq:cost}
\end{equation}
where $r_k=(i_1,\dots,i_{L_k})$ is the customer sequence of route $k$.

The solution can equivalently be written as a depot-separated sequence, e.g.\ $(0,3,5,2,0,4,1,6,0)$, so that CVRP becomes a sequential decision process: a policy selects one node at a time, the remaining capacity decreases with each customer and resets at each depot visit. This capacity-masked formulation is the one used by the attention-model policy and implemented in the RL4CO environment (Section~\ref{sec:setup}).

\subsection{Training objective}
Following standard practice for attention-model routing policies, the policy is trained with REINFORCE and a rollout baseline~\cite{sutton1999,kool2019}. For a problem instance $x$ sampled from a distribution $P(x)$, and a policy $\pi_\theta$ that samples a solution $a$, the reward is the negative of the cumulative length of all routes,
\begin{equation}
R(a,x) = -C(S) = -\sum_{k=1}^{m} C(r_k),
\label{eq:reward}
\end{equation}
where $S=\{r_1,\dots,r_m\}$ is the solution encoded by the action sequence $a$ and $C(r_k)$ the length of route $r_k$ from Eq.~\eqref{eq:cost}. The gradient of the training objective, which is maximised, is
\begin{equation}
\nabla_\theta \mathcal{L}(\theta \mid x) = \mathbb{E}_{\pi_\theta(a\mid x)}\big[(R(a,x) - b(x))\,\nabla_\theta \log \pi_\theta(a\mid x)\big],
\label{eq:reinforce}
\end{equation}
where $b(x)$ is a baseline (the reward obtained by a frozen, periodically-updated copy of the best-performing policy, decoded greedily) used to reduce gradient variance. During training, actions are sampled from the categorical distribution over feasible nodes to encourage exploration; during validation and testing, actions are selected greedily.

\subsection{Reference policy: the attention model}
\label{sec:reference}
As the policy for this decision process and the reference point for the head-to-head comparison, we adopt the attention-model policy of Kool et al.~\cite{kool2019}, trained with Eq.~\eqref{eq:reinforce}. The encoder first embeds depot and customer nodes with two separate linear layers of equal output dimension $D$ (customers carry three raw features: coordinates and demand; the depot carries two), then processes the concatenated embeddings through $L$ MHA layers (Eqs.~\eqref{eq:attention} and~\eqref{eq:mha}) with residual connections~\cite{he2016}, batch normalization, and a feed-forward sub-layer, producing final node embeddings $H=[h_0,h_1,\dots,h_n]$.

The decoder pre-computes, once per instance, a graph-level context $g$ and glimpse key/value/logit projections from $H$. The reference graph context uses mean pooling over node embeddings followed by a linear projection,
\begin{equation}
g = W_g\,\bar{h} + b_g, \qquad \bar h = \frac1N\sum_i h_i,
\label{eq:meanctx}
\end{equation}
and at each decoding step
$t$, a dynamic step context concatenates the embedding of the previously visited node with the vehicle's remaining capacity $c_t$, $s_t = [h_{a_{t-1}} \| c_t]$, projects it linearly, and adds it to the static graph context to form the glimpse query $q^{glimpse}_t = g + W_s s_t + b_s$. This query attends (via Eq.~\eqref{eq:attention}, with a mask over infeasible nodes) over the pre-computed glimpse keys/values to produce logits $u^t_i$ over the candidate nodes (Eq.~\eqref{eq:pointer}), from which the next node is sampled (training) or chosen greedily (evaluation).

\section{Sources of Degradation and Approaches}
\label{sec:method}

\begin{figure*}[ht]
\centering
\includegraphics[width=\linewidth]{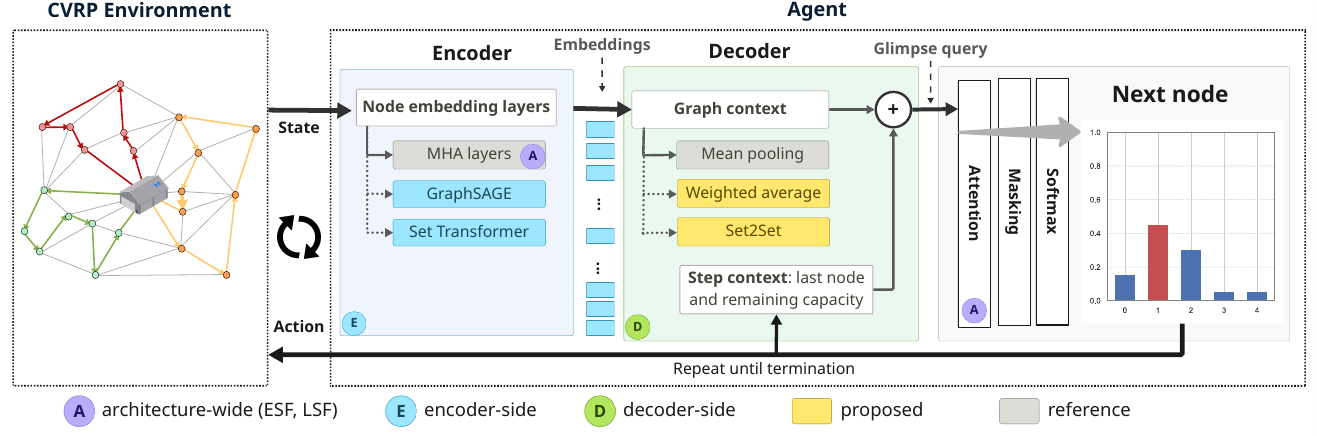}
\caption{Overview of the encoder-decoder attention-model policy: the reference architecture together with the architecture-wide, encoder-side and decoder-side approaches evaluated in this work.}
\label{fig:overview}
\end{figure*}

Fig.~\ref{fig:overview} summarises the policy and the locations of the evaluated approaches. A CVRP instance is given as a graph of the depot and the clients with their coordinates and demands. The encoder maps every node of this graph to a latent representation, the node embedding. From these embeddings the decoder forms a context that combines a summary of the whole graph with the current state of the vehicle. At each step, this context is compared with the embeddings of all feasible nodes to obtain one score per node, and a softmax over these scores gives the probability of each node to be visited next, from which the next client or a return to the depot is selected. The approaches studied in this paper modify this pipeline at three levels: architecture-wide in the attention of encoder and decoder (Section~\ref{sec:method-global}), in the encoder that produces the embeddings (Section~\ref{sec:method-encoder}), and in the decoder aggregation that forms the graph context (Section~\ref{sec:method-context}).

\subsection{Sources of cross-size degradation}
\label{sec:method-problems}
The reference policy (Section~\ref{sec:reference}) contains several components whose behaviour depends on the number of nodes. From its structure and from the literature we derive three candidate sources of degradation, which define the levers studied in this paper; Section~\ref{sec:res-sources} confirms them empirically.

The first source, S1 (node density), describes \emph{what} changes between sizes. With fixed capacity and demand range, a larger instance places more and closer nodes in the same area and needs more tours, so a policy trained on one size meets a different density at test time. The other two sources are mechanisms inside the policy that react to this change, and the approaches in this paper act on these two mechanisms.
S2 (attention dispersion): the softmax in Eqs.~\eqref{eq:attention} and~\eqref{eq:pointer} normalises over all nodes, so attention becomes flatter as $N$ grows~\cite{wang2025}. 
S3 (graph-context dilution): the mean-pooled context (Eq.~\eqref{eq:meanctx}) weights every node equally, so with more nodes it moves towards the same average vector and carries less instance-specific information; since the decoder reuses it at every step, this loss affects every decision.

Table~\ref{tab:levers} assigns each approach to the source it addresses and to its origin: approaches marked as \emph{proposed} are contributions of this paper, whereas the cited approaches are taken from the literature and re-implemented in one common framework. Each changes only the component under test; everything else is identical to the reference. The selector (Section~\ref{sec:method-selector}) then combines the trained approaches.

\begin{table}[t]
\centering
\caption{Approaches, the degradation source they address and their origin. A: architecture-wide, E: encoder-side, D: decoder-side. Origin: citation, taken from the literature; proposed, contribution of this paper.}
\label{tab:levers}
\small\setlength{\tabcolsep}{3pt}
\begin{tabular}{llll}
\toprule
 & Approach & Source & Origin \\
\midrule
A & Entropy-based scaling (ESF) & S2 & \cite{xiao2025} \\
A & Learnable scaling (LSF) & S2 & \cite{xiao2025} (suggested) \\
E & GraphSAGE encoder & S2 & \cite{hamilton2017} \\
E & Set Transformer encoder & S2, S3 & \cite{lee2019} \\
D & Weighted-average context & S3 & proposed \\
D & Set2Set context & S3 & proposed \\
D & GRU context & S3 & proposed \\
\midrule
 & Approach selector & all & proposed \\
\bottomrule
\end{tabular}
\end{table}

\subsection{Architecture-wide approaches: size-aware attention}
\label{sec:method-global}
Attention dispersion (S2) is the most direct consequence of a change in $N$, since it acts on the attention scores themselves. Rescaling these scores with the graph size is therefore the most targeted lever, and it changes no learned weights.

\paragraph{Entropy-based scaling factor (ESF)}
We adopt ESF (Eq.~\eqref{eq:esf})~\cite{xiao2025} because it is established for this problem and needs no retraining. We apply the factor in the scaled dot-product attention of both encoder and decoder of the trained reference models. For $n>n_{\mathrm{train}}$ the factor exceeds one and sharpens the softmax, which counteracts the flattening caused by the larger number of keys (S2); for $n<n_{\mathrm{train}}$ it softens the softmax accordingly.

\paragraph{Learnable scaling factor (LSF)}
ESF corrects dispersion (S2) with a fixed logarithmic rule, which need not be the best sharpening, and it ties the model to one training size. A learnable factor, suggested as an extension of ESF~\cite{xiao2025}, lets the policy learn this correction from data and allows training on several sizes. The factor in Eq.~\eqref{eq:esf} is replaced by a learned function of the instance size, $\lambda_n = \exp\!\big(f_\phi(\log n)\big)$, where $f_\phi$ is a small feed-forward network (hidden dimension 16) with the scalar input $\log n$; the exponential keeps the factor positive. The unit is trained jointly with the policy under the same REINFORCE objective (Eq.~\eqref{eq:reinforce}), with no additional loss term, and is applied at the same locations as ESF. Because single-size training shows the unit only one value of $\log n$, LSF is additionally trained sequentially on two sizes (Section~\ref{sec:res-seq}).

\subsection{Encoder-side approaches}
\label{sec:method-encoder}
The encoder produces the node embeddings from which both the graph context and the pointer keys are computed, so a size-robust encoder addresses dispersion (S2), and possibly context dilution (S3), at their source. In full self-attention, each node mixes information from all $N$ nodes through a softmax whose sharpness depends on $N$. We therefore test two established encoders that remove this dependence in different ways: by replacing the softmax with a uniform mean over the neighbourhood, and by routing all interaction through a fixed-width bottleneck. Both replace the MHA encoder while the decoder, the node-embedding layers and the training procedure stay identical.

\paragraph{GraphSAGE encoder}
GraphSAGE is designed to generalize to graphs unseen during training~\cite{hamilton2017}, which matches the size-transfer setting. We replace the MHA encoder by three GraphSAGE layers with the mean aggregator (Eq.~\eqref{eq:sage}). Unlike the original method, no neighbourhood is sampled: $N(v)$ contains all other nodes of the instance. The message to node $v$ is then $(\sum_u h_u - h_v)/(|V|-1)$, so, in contrast to self-attention, all other nodes are weighted equally and the aggregate is normalised by the neighbourhood size, which keeps its scale independent of $N$. The encoder thus avoids the size-dependent softmax over all nodes (S2). Its message is, however, itself a mean over all nodes and therefore subject to the same dilution as the graph context (S3).

\paragraph{Set Transformer encoder}
The Set Transformer~\cite{lee2019} is designed for unordered sets and thus matches the permutation-invariant node set of a CVRP instance. We replace the MHA encoder by the Set Transformer encoder with three ISAB layers (Eq.~\eqref{eq:isab}), $m=16$ inducing points shared across instances and 8 heads in both MABs, and keep the reference decoder. Each node then receives information through the same $m$ summary vectors for any $N$, which limits how strongly its embedding, and the context and keys derived from it, depend on the graph size (S2, S3).

\subsection{Proposed: learned graph-context aggregation}
\label{sec:method-context}
The graph context is the only summary of the whole instance that the decoder receives, and it enters the glimpse query at every decoding step, so its dilution (S3) affects every decision. We therefore propose to replace the uniform mean of Eq.~\eqref{eq:meanctx} by a learned aggregation that can decide which nodes shape the summary. Two design requirements follow from the problem. First, the context is computed before any route exists, so the node set has no meaningful order and the aggregation must be permutation-invariant. Second, it must accept any number of nodes and return a fixed-size vector. The weighted average and Set2Set satisfy both requirements, whereas the GRU deliberately drops the first one to test whether permutation invariance matters. All three add few parameters, and leave the encoder, the step context and the pointer unchanged; in RL4CO the graph context is computed once per instance in the decoder module.

\paragraph{Weighted-average graph context} The simplest learned extension of the mean gives each node a learned weight, \begin{equation} \bar{h}_w = \sum_i w_i h_i, \qquad w_i = \mathrm{softmax}_i(s_i), \label{eq:wavg} \end{equation} which replaces $\bar h$ in Eq.~\eqref{eq:meanctx}. The scores $s_i$ come from a two-layer scoring network ($128\to128\to1$) with a ReLU after the first layer. Because the softmax normalises over the nodes, the context stays a convex combination of embeddings for any $N$, but it is no longer uniform. To test whether the gain depends on what the scoring network can express, a reduced variant with a single linear scoring layer, $s_i = W h_i + b$, is evaluated as an ablation.

\paragraph{Set2Set graph context} A single weighting pass can capture only one view of the instance. We therefore use the Set2Set readout (Eq.~\eqref{eq:set2set})~\cite{vinyals2016} over the node embeddings as the graph context of the attention-model decoder, with $T=3$ processing steps. Its repeated attention lets the summary capture interactions between nodes beyond a single mean, and it is order-invariant by construction.

\paragraph{GRU graph context} To test whether permutation invariance matters, we also propose an order-dependent aggregation: the node embeddings are passed in sequence through a GRU and its last hidden state is the graph context. It also gates how much each node contributes, but its result depends on the node order.

\subsection{Proposed: approach selector}
\label{sec:method-selector}
The levers above act on different sources of degradation, so there is no reason to expect a single one to be best for every transfer; Section~\ref{sec:results} confirms that upscaling and downscaling respond to different levers. Rather than forcing one architecture, the selector keeps all trained models and chooses one per situation. This matches the operational setting: on a given day the number of clients $n$ is known when the instance is formed, and the training size $a$ of the available models is known. For each approach $k$ in the set $\mathcal{K}$ of trained approaches and each training size $a$, a quadratic function of the problem size is fitted through its average costs measured on the CVRP20, CVRP50 and CVRP100 test sets,
\begin{equation}
\hat z_{k,a}(n) = \alpha_{k,a} n^2 + \beta_{k,a} n + \gamma_{k,a},
\label{eq:selector-fit}
\end{equation}
and the selector solves the instance with the model of the approach with the lowest estimated cost,
\begin{equation}
k^*(a,n) = \arg\min_{k\in\mathcal{K}} \hat z_{k,a}(n).
\label{eq:selector}
\end{equation}
On the three measured sizes, Eq.~\eqref{eq:selector-fit} reproduces the measured costs; between them it interpolates, and beyond CVRP100 it extrapolates. The selector needs no additional training and is evaluated on a fresh instance set (Section~\ref{sec:setup}).

\section{Experimental Setup}
\label{sec:setup}

\subsection{Framework, hardware and environment}
All experiments use the RL4CO framework~\cite{berto2023}, which provides the CVRP environment, the attention-model policy, the REINFORCE-with-rollout training loop and evaluation utilities shared by the reference and all variants. All models were trained on one NVIDIA RTX~3070~Ti GPU with PyTorch~2.11 and PyTorch Geometric~2.7.

\subsection{Dataset generation}
CVRP instances are generated with the framework's CVRP generator. Validation and test sets contain 10{,}000 instances each, generated once with fixed seeds (4321 and 1234), so every model is scored on identical instances. Training instances are resampled on the fly (100{,}000 per epoch) from the same distribution to avoid a static-dataset bias. Node coordinates are uniform in the unit square, the capacity is fixed at $Q=30$ and integer demands are drawn from $\{1,\dots,9\}$ for all sizes, so CVRP20, CVRP50 and CVRP100 differ only in the number of clients. Unlike the setting of Kool et al.~\cite{kool2019}, where capacity grows with the problem size, this isolates the effect of the number of nodes. Because the selector is built from the costs on the test set, it is evaluated on a separate instance set generated with seed 6249, for CVRP20, CVRP50 and CVRP100 and for the unseen sizes CVRP35, CVRP75 and CVRP200.

\subsection{Models and training procedure}
Every trained approach is trained independently and from scratch on each of the three sizes (the sequential training in Section~\ref{sec:res-seq} is the only exception). ESF needs no training and is applied at inference time to the three trained reference models. Training follows the REINFORCE-with-rollout procedure of Section~\ref{sec:formulation}(Eq.~\eqref{eq:reinforce}): actions are sampled during training and decoded greedily during validation and testing, and the rollout baseline is replaced by the current policy when it outperforms the frozen baseline on a held-out set. All approaches share the same hyperparameters: embedding dimension 128, feed-forward width 512, 8 attention heads, 3 encoder layers, temperature 1.0, tanh clipping 10.0, Adam with learning rate $10^{-4}$, batch size 512 and 150 training epochs. Each change is confined to the component under test and no variant receives a larger training budget.

\subsection{Evaluation methodology}
For every (training size, test size) pair, the average solution cost (total route length, the negative of the reward in Eq.~\eqref{eq:reward}) is computed over the full test set. To measure generalization independently of absolute solution quality, the cost of a model trained on size $a$ and evaluated on size $b$ is compared with the cost of the model of the same approach trained natively on size $b$:
\begin{equation}
\mathrm{Gap}_{a\to b}(\%) = \frac{z_{a\to b} - z_{b\to b}}{z_{b\to b}}\times 100,
\label{eq:gap}
\end{equation}
where $z_{a\to b}$ is the average cost of the model trained on $a$ and tested on $b$, and $z_{b\to b}$ that of the model trained and tested on $b$. A smaller gap indicates better generalization; it is evaluated in both directions, upscaling ($a<b$) and downscaling ($a>b$).

Each approach is then compared head-to-head with the reference by its \emph{gap gain},
\begin{equation}
\Delta_{a\to b} = \mathrm{Gap}^{\mathrm{ref}}_{a\to b} - \mathrm{Gap}_{a\to b} \quad \text{(percentage points)},
\label{eq:gain}
\end{equation}
so that a positive $\Delta$ means the approach generalizes better than the reference on that transfer. Because the denominator in Eq.~\eqref{eq:gap} is the approach's own native model, the gap gain is read together with the same-size costs (Table~\ref{tab:native}), and absolute costs are reported where the distinction matters.

\section{Results and Discussion}
\label{sec:results}

The reference is reported with costs and gaps (Eq.~\eqref{eq:gap}, Table~\ref{tab:baseline}), the approaches by their gap gain (Eq.~\eqref{eq:gain}, Table~\ref{tab:gain}) in percentage points (pp) on the three upscaling and three downscaling transfers. After the reference and its diagnosis, the approaches follow the order of Section~\ref{sec:method}.

\subsection{Generalization loss of the reference policy}
\label{sec:res-reference}

Table~\ref{tab:baseline} reports the reference attention-model policy of Kool et al.~\cite{kool2019} as implemented in RL4CO~\cite{berto2023}, an established, already-tuned construction policy, so every improvement below is a gain over a strong reference. Its cross-size behaviour reproduces the bidirectional bottleneck. First, the gap grows with the size mismatch: a one-step transfer costs $0.55$--$2.61\%$, whereas the two-step transfers cost $5.23\%$ (CVRP20$\to$CVRP100) and $7.77\%$ (CVRP100$\to$CVRP20). Second, the gap is \emph{asymmetric}: for every size pair, downscaling is worse than upscaling (e.g.\ CVRP100$\to$CVRP20 $7.77\%$ vs.\ CVRP20$\to$CVRP100 $5.23\%$). A policy trained on sparse graphs therefore transfers to dense graphs more readily than the reverse.

\begin{table}[h]
\centering
\caption{Reference attention-model policy: average solution cost and, in
brackets, cross-size optimality gap (\%). Rows are training size, columns
are test size.}
\label{tab:baseline}
\small\setlength{\tabcolsep}{4pt}
\begin{tabular}{lccc}
\toprule
Train & CVRP20 & CVRP50 & CVRP100 \\
\midrule
CVRP20  & 6.480 (0.00) & 13.692 (2.18) & 25.408 (5.23) \\
CVRP50  & 6.649 (2.61) & 13.400 (0.00) & 24.277 (0.55) \\
CVRP100 & 6.984 (7.77) & 13.611 (1.58) & 24.145 (0.00) \\
\bottomrule
\end{tabular}
\end{table}

\subsection{Sources of cross-size degradation}
\label{sec:res-sources}

The measurements below confirm the three sources of Section~\ref{sec:method-problems} on the reference policy.

\paragraph{S1: the distribution shift is a shift in node density}
On the test sets, the mean normalised demand per client ($0.167$), the mean depot-to-client distance ($0.52$) and the mean pairwise distance ($0.52$) are practically constant across sizes, while the mean nearest-neighbour distance shrinks from $0.124$ (CVRP20) through $0.075$ (CVRP50) to $0.052$ (CVRP100). The mean total demand per instance, relative to the capacity and thus a lower bound on the average number of routes, grows from $3.34$ through $8.33$ to $16.68$. The generalization gap studied here is therefore a response to node \emph{density} and to the number of required tours, not to a change of operating region or demand regime.

\begin{figure}[h]
\centering
\includegraphics[width=\columnwidth]{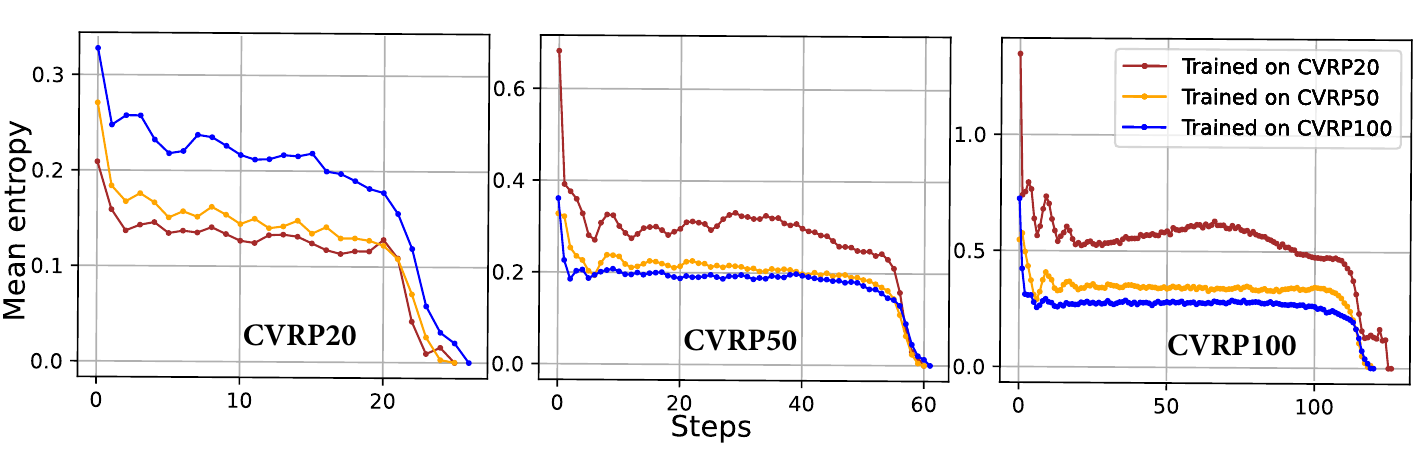}
\caption{Entropy of the decoder action distribution of the reference policy at each decoding step, averaged over the test instances. Panels from left to right: test on CVRP20, CVRP50 and CVRP100 (note the different vertical scales); lines: models trained on CVRP20, CVRP50 and CVRP100.}
\label{fig:diag}
\end{figure}

\paragraph{S2: attention dispersion}
Attention dispersion is measured by the entropy of the decoder's action distribution, $H(p_t)$, which is high when the probability mass is spread over many candidate nodes and low when the policy points clearly to one node. Fig.~\ref{fig:diag} shows this entropy at every decoding step for the three reference models on each test size. Two patterns are visible. First, within each panel the entropy grows with the mismatch between training and test size: on CVRP100 the CVRP20-trained model stays roughly twice as uncertain as the CVRP100-trained model over the whole route, and on CVRP20 the order is reversed, with the CVRP100-trained model highest. Second, every curve peaks at the first decoding step, when all customers are still candidates, and falls to zero at the end of the route, when few feasible nodes remain. A policy applied to an unfamiliar size therefore separates good from poor candidates less sharply, in line with Wang et al.~\cite{wang2025}. On both extremes the higher entropy comes with the higher cost (CVRP20: $6.984$ for the CVRP100-trained vs.\ $6.480$ for the CVRP20-trained model; CVRP100: $25.408$ vs.\ $24.145$). The relation is not strictly monotone (on CVRP50, the CVRP100-trained model has slightly lower entropy than the CVRP50-trained model but produces longer routes), which indicates that dispersion is one source among several.

\begin{figure}[h]
\centering
\includegraphics[width=\columnwidth]{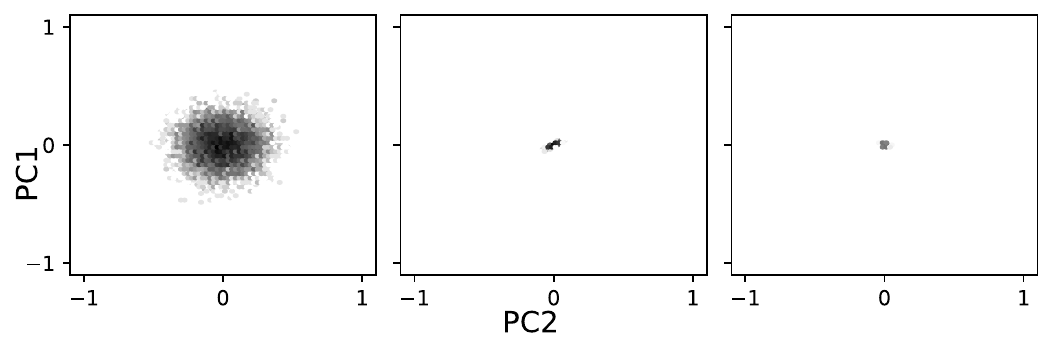}
\caption{Mean-pooled graph context of the CVRP100-trained reference policy for the individual test instances, projected on its first two principal components. Panels from left to right: test on CVRP20, CVRP50 and CVRP100; darker cells contain more instances.}
\label{fig:context}
\end{figure}

\paragraph{S3: graph-context dilution}
Fig.~\ref{fig:context} visualises the graph context that the CVRP100-trained model computes for each test instance. On CVRP20 the contexts of different instances spread over a wide region, on CVRP50 they shrink to a small cluster, and on CVRP100 they collapse almost to a single point: averaging over more nodes removes the differences between instances, so the decoder receives nearly the same summary for every large instance. Quantitatively, the variance of the context across instances falls for all three reference models from about $1.4\times10^{-3}$ on CVRP20 to $3$--$4\times10^{-4}$ on CVRP100, and the learned projection compresses it further (to $8$--$9\times10^{-5}$ on CVRP100). The figure also links dilution to the downscaling asymmetry of Table~\ref{tab:baseline}: a model trained on CVRP100 has only seen this narrow cluster of contexts, whereas on CVRP20 it receives contexts far outside that range. Because the decoder reuses the context at every step, this mismatch affects every decision, which motivates a richer, learned aggregation.

\subsection{Architecture-wide approaches: size-aware attention}
\label{sec:res-global}

\begin{table}[h]
\centering
\caption{Gap gain $\Delta$ of every approach with respect to the reference (pp, Eq.~\eqref{eq:gain}). \textcolor{blue}{$\uparrow$}: smaller gap than the reference (improvement), \textcolor{red}{$\downarrow$}: larger gap (degradation). A: architecture-wide, E: encoder-side, D: decoder-side; $^\dagger$ proposed. Best gain per transfer in bold. The gains of GraphSAGE are read together with its weaker same-size models (Table~\ref{tab:native}).}
\label{tab:gain}
\resizebox{\columnwidth}{!}{%
\begin{tabular}{llrrrrrr}
\toprule
 & & \multicolumn{3}{c}{Upscaling} & \multicolumn{3}{c}{Downscaling} \\
\cmidrule(lr){3-5}\cmidrule(lr){6-8}
 & Approach & 20$\to$50 & 20$\to$100 & 50$\to$100 & 50$\to$20 & 100$\to$20 & 100$\to$50 \\
\midrule
A & ESF                  & \gup{0.48} & \gup{1.41} & \gup{0.03} & \gup{\textbf{0.37}} & \gup{0.24} & \gdn{0.04} \\
A & LSF                  & \gup{0.23} & \gup{0.34} & \gup{0.06} & \gdn{0.16} & 0.00 & \gup{0.43} \\
E & GraphSAGE            & \gup{\textbf{1.43}} & \gup{\textbf{3.07}} & \gup{\textbf{1.84}} & \gdn{2.07} & \gdn{5.42} & \gdn{1.78} \\
E & Set Transformer      & \gup{0.67} & \gup{0.49} & \gdn{0.15} & \gdn{0.31} & \gdn{1.08} & \gup{\textbf{0.54}} \\
D & Weighted avg.$^\dagger$ & \gup{0.16} & \gdn{0.41} & \gdn{0.09} & \gdn{0.09} & \gup{0.91} & \gup{0.33} \\
D & Set2Set$^\dagger$    & \gdn{0.36} & \gdn{0.60} & \gup{0.03} & \gdn{0.06} & \gup{\textbf{0.92}} & \gup{0.25} \\
D & GRU$^\dagger$         & \gdn{0.28} & \gdn{0.93} & \gdn{0.19} & \gdn{1.16} & \gdn{0.08} & \gup{0.27} \\
\bottomrule
\end{tabular}}
\end{table}

\paragraph{Entropy-based scaling factor (ESF)}
Table~\ref{tab:gain} lists the gap gains of all approaches. Because $\lambda_n=1$ on the training size, ESF leaves same-size behaviour unchanged, so its gains are purely cross-size. It is positive on five of the six transfers and gives a strong improvement on the hardest upscaling transfer: CVRP20$\to$CVRP100 gains $+1.41$\,pp (from $5.23\%$ to $3.83\%$) and CVRP20$\to$CVRP50 gains $+0.48$\,pp. It also gives the best gain on CVRP50$\to$CVRP20 ($+0.37$\,pp) and $+0.24$\,pp on CVRP100$\to$CVRP20, with only CVRP100$\to$CVRP50 marginally negative ($-0.04$\,pp). ESF was originally evaluated with capacity growing with the problem size; these results show that it remains effective when only the number of nodes changes, without any retraining.

\paragraph{Learnable scaling factor (LSF)}
LSF is positive on four transfers and gives $+0.43$\,pp on CVRP100$\to$CVRP50. In addition, it lowers the same-size cost on all three sizes (e.g.\ $24.145\to24.116$ on CVRP100) and the cost in eight of the nine train--test cells, so its gap gains, measured against its own better native models, understate its absolute improvement. Its cross-size gains are smaller than those of ESF (e.g.\ $+0.34$ vs.\ $+1.41$\,pp on CVRP20$\to$CVRP100) because, trained on a single size, the network observes only one value of $\log n$, whereas ESF encodes the size dependence analytically. The sequential training below shows that a second training size releases this potential.

\paragraph{LSF with sequential two-size training}\label{sec:res-seq}
Because single-size training exposes LSF to a single value of $\log n$, we also trained it sequentially (75 epochs per size): an upscaling model on CVRP20 then CVRP50, tested on the unseen CVRP100, and a downscaling model on CVRP100 then CVRP50, tested on the unseen CVRP20. The reference trained in the same way has gaps of $3.73\%$ (upscaling) and $5.65\%$ (downscaling). LSF reaches $2.18\%$ ($+1.55$\,pp), the largest upscaling gain of the sequential setting, and $6.17\%$ ($-0.52$\,pp) in downscaling. A second training size thus gives LSF a large upscaling gain, which makes it a promising candidate for true mixed-size training.

\subsection{Encoder-side approaches}
\label{sec:res-encoder}

\paragraph{GraphSAGE encoder}
GraphSAGE gives the largest gap gains in Table~\ref{tab:gain} on all three upscaling transfers ($+1.43$, $+3.07$ and $+1.84$\,pp). The CVRP20$\to$CVRP100 gap falls from $5.23\%$ to $2.17\%$, and on CVRP50$\to$CVRP100 the gap becomes negative ($-1.29\%$): the CVRP50-trained model beats the CVRP100-trained one on CVRP100. This indicates that the normalised, content-independent message lets embeddings learned on smaller graphs carry over to larger ones. Its same-size costs are $2.4$ to $4.2\%$ above the reference (Table~\ref{tab:native}), so these gains reflect a flatter cost profile across sizes rather than cheaper transfers in absolute terms (e.g.\ $25.705$ vs.\ $25.408$ on CVRP20$\to$CVRP100). In the downscaling direction, all three transfers get worse ($-2.07$, $-5.42$ and $-1.78$\,pp), since relationships learned by aggregating over many nodes need more nodes than a small graph provides.

\paragraph{Set Transformer encoder}
The Set Transformer improves both transfers from CVRP20 ($+0.67$ and $+0.49$\,pp) and gives the best gain on CVRP100$\to$CVRP50 ($+0.54$\,pp), while keeping its same-size costs within $0.2\%$ of the reference (Table~\ref{tab:native}). Its upscaling gains therefore translate into cheaper solutions: CVRP20$\to$CVRP50 falls from $13.692$ to $13.627$ and CVRP20$\to$CVRP100 from $25.408$ to $25.258$. This supports the fixed-width bottleneck as an upscaling lever that preserves same-size quality. The remaining transfers are negative, most strongly CVRP100$\to$CVRP20 ($-1.08$\,pp), so, as for GraphSAGE, the benefit is directional.

\subsection{Proposed: learned graph-context aggregation}
\label{sec:res-context}

\paragraph{Weighted-average graph context}
In Table~\ref{tab:gain}, the weighted average gives a large gain on the hardest downscaling transfer (CVRP100$\to$CVRP20, $+0.91$\,pp) together with $+0.33$\,pp on CVRP100$\to$CVRP50 and $+0.16$\,pp on CVRP20$\to$CVRP50. Unlike GraphSAGE, it achieves this while keeping same-size quality (e.g.\ $24.069$ vs.\ $24.145$ on CVRP100), so the gain is a real cost reduction: on CVRP100$\to$CVRP20 the average cost falls from $6.984$ to about $6.91$. It has a lower cost than the reference in seven of the nine train--test cells, and its losses on the remaining transfers are moderate ($-0.09$ to $-0.41$\,pp). The ablation with a single linear scoring layer loses the downscaling benefit (CVRP100$\to$CVRP20: $-1.30$\,pp), which indicates that the improvement comes from the expressiveness of the scoring network and not merely from abandoning the uniform mean.

\paragraph{Set2Set graph context}
Set2Set produces the largest single downscaling gain in the study, $+0.92$\,pp on CVRP100$\to$CVRP20 (from $7.77\%$ to $6.86\%$, an average cost of about $6.93$), together with $+0.25$\,pp on CVRP100$\to$CVRP50, and it has the lowest same-size cost on CVRP50 ($13.374$) and the second-lowest on CVRP100 ($24.050$). Its gains are concentrated on transfers from the dense regime, where the dilution of the mean was measured (Section~\ref{sec:res-sources}): a richer summary learned on dense graphs remains informative on sparse ones. Transfers from CVRP20 do not benefit ($-0.36$ and $-0.60$\,pp), which is consistent with the measurements: on sparse graphs the mean is less diluted, so a learned summary has less to correct.

\paragraph{GRU graph context}
The order-dependent GRU improves only CVRP100$\to$CVRP50 ($+0.27$\,pp) and worsens the other five transfers ($-0.08$ to $-1.16$\,pp), although it has the lowest same-size cost on CVRP100 ($24.030$). A good same-size summary is therefore not sufficient: the GRU can learn order artefacts that do not transfer across sizes, which supports permutation invariance as a design requirement of the proposed aggregations.

\subsection{Proposed: approach selector}
\label{sec:res-selector}

\begin{table}[h]
\centering
\caption{Approach selector on the fresh instance set (seed 6249): cost reduction relative to the reference (\%), $(z^{\mathrm{ref}}-z^{\mathrm{sel}})/z^{\mathrm{ref}}\times100$; \textcolor{blue}{$\uparrow$}: cheaper than the reference, \textcolor{red}{$\downarrow$}: more expensive. CVRP35, CVRP75 and CVRP200 are unseen sizes.}
\label{tab:selector}
\small\setlength{\tabcolsep}{3.5pt}
\begin{tabular}{lcccccc}
\toprule
 & \multicolumn{3}{c}{Seen sizes} & \multicolumn{3}{c}{Unseen sizes} \\
\cmidrule(lr){2-4}\cmidrule(lr){5-7}
Train & 20 & 50 & 100 & 35 & 75 & 200 \\
\midrule
CVRP20  & \gup{0.20} & \gup{0.47} & \gup{0.56} & \gup{0.26} & \gup{0.49} & \gup{3.20} \\
CVRP50  & \gup{0.08} & \gup{0.25} & \gup{0.40} & \gdn{0.03} & \gup{0.35} & \gup{0.46} \\
CVRP100 & \gup{1.07} & \gup{0.41} & \gup{0.43} & \gup{0.49} & \gup{0.41} & \gup{0.35} \\
\bottomrule
\end{tabular}
\end{table}

Table~\ref{tab:selector} shows that the selector turns the complementarity of the levers into a consistent improvement. On the fresh instance set it lowers the cost of the reference in all nine seen train--test pairs, by up to $1.07\%$ on CVRP100$\to$CVRP20, the hardest downscaling transfer. It also lowers the cross-size gap in four of the six transfers, by $+0.14$ to $+0.95$\,pp (CVRP50$\to$CVRP20 and CVRP50$\to$CVRP100 change by $-0.14$ and $-0.03$\,pp, since a lower native cost raises the gap of an otherwise cheaper transfer). On the unseen sizes it is cheaper in eight of nine pairs, and the largest gain occurs for the largest extrapolation: the CVRP20-trained models reach $48.43$ instead of $50.04$ on CVRP200 ($3.2\%$). The only exception, CVRP50$\to$CVRP35, differs by $0.03\%$.

\subsection{Synthesis: different levers for upscaling and downscaling}
\label{sec:res-synthesis}

The comparison shows that the two transfer directions respond to different levers (Table~\ref{tab:gain}). \emph{Upscaling} responds to the encoder approaches and to attention rescaling. Among the approaches that keep same-size quality, ESF is the strongest on the hardest upscaling transfer: it reaches $+1.41$\,pp on CVRP20$\to$CVRP100 without retraining and gives the lowest cost of all approaches there ($25.068$ vs.\ $25.408$ for the reference), while the Set Transformer is strongest on CVRP20$\to$CVRP50 ($+0.67$\,pp). \emph{Downscaling}, in particular the hardest CVRP100$\to$CVRP20 transfer, responds most to the proposed graph-context aggregation: Set2Set and the weighted average each recover about $0.9$\,pp, whereas the encoder approaches make it worse. The effective downscaling levers share one mechanism, a more expressive summary of a dense graph that remains informative on a sparse one. This matches the asymmetry of the reference: the encoder and attention levers help in the direction in which density increases, and the context levers in the direction in which the training-size context was diluted.

\begin{table}[h]
\centering
\caption{Same-size average cost (train = test) of the reference and of every
approach; lower is better. ESF equals the reference by construction.}
\label{tab:native}
\small
\begin{tabular}{lccc}
\toprule
Approach & CVRP20 & CVRP50 & CVRP100 \\
\midrule
Reference / ESF      & 6.480 & 13.400 & 24.145 \\
LSF                  & 6.475 & 13.393 & 24.116 \\
GraphSAGE            & 6.634 & 13.793 & 25.160 \\
Set Transformer      & 6.474 & 13.423 & 24.115 \\
Weighted avg.$^\dagger$ & 6.466 & 13.408 & 24.069 \\
Set2Set$^\dagger$    & 6.487 & 13.374 & 24.050 \\
GRU$^\dagger$         & 6.476 & 13.406 & 24.030 \\
\bottomrule
\end{tabular}
\end{table}

Most approaches keep the same-size quality (Table~\ref{tab:native}): they stay within $0.5\%$ of the reference cost on every size, except GraphSAGE ($+2.4$, $+2.9$ and $+4.2\%$). Same-size quality and cross-size gap are therefore separate objectives, and the selector chooses on estimated cost.


\section{Conclusion and Future Work}
\label{sec:conclusion}

This paper started from a practical requirement for delivery route planning: a routing policy must handle a variable, a priori unknown number of clients. Using CVRP with fixed vehicle capacity, in which only the number of clients changes, we identified three sources of the cross-size degradation of the attention-based Pointer Network (node density, attention dispersion and graph-context dilution) and compared levers that act on the two policy-internal sources at three levels of its encoder--decoder architecture: architecture-wide, encoder-side and decoder-side.

The reference gap grows with the size mismatch and is asymmetric: downscaling (CVRP100 to CVRP20, $7.77\%$) is harder than upscaling (CVRP20 to CVRP100, $5.23\%$). The comparison shows that the two directions respond to different levers. For upscaling, ESF lowers the CVRP20 to CVRP100 gap by $1.41$ percentage points without retraining and gives the lowest cost there, and the Set Transformer encoder improves the transfers from CVRP20 with same-size quality preserved. For downscaling, the proposed learned graph-context aggregations give the largest gains on the hardest transfer: Set2Set and the weighted average lower CVRP100 to CVRP20 by $0.92$ and $0.91$ percentage points while keeping same-size quality, and the order-dependent GRU context supports permutation invariance as a requirement. With a second training size, the learnable scaling factor gains $1.55$ points in upscaling. The proposed approach selector exploits this complementarity with a data-driven decision on estimated cost rather than a hand-crafted rule: it lowers the cost of the reference in all nine seen train--test pairs and in eight of nine pairs on the unseen sizes CVRP35, CVRP75 and CVRP200, by up to $3.2\%$.

\bigskip
Three future directions follow from these findings. First, the strongest upscaling and downscaling levers act on different components, so attention rescaling and the proposed graph-context aggregation can be combined in one model and trained on several sizes at once, building on the sequential results of the learnable scaling factor. Second, the selector currently decides on problem size only; an ensemble that keeps the cheapest of several solutions, or a small learned selection network that also uses instance features such as demand relative to capacity or client density, can choose per instance. Third, the comparison will be repeated on larger instances.

\section*{Funding}
The research project was funded by The Ministry of
the Environment, Nature Conservation, and Transport of the
State of North Rhine-Westphalia in Germany, and co-financed
by the European Union for the research project ``AI4DRONE'', with grant
number IN-NX-1-033b.

\bibliographystyle{elsarticle-num}
\bibliography{References}

\end{document}